\documentclass[conference]{IEEEtran}
\IEEEoverridecommandlockouts
\usepackage{cite}
\usepackage{amsmath,amssymb,amsfonts}
\usepackage{graphicx}
\usepackage{enumitem}
\usepackage{textcomp}
\usepackage{algorithm}
\usepackage{algorithmic}
\usepackage[algo2e]{algorithm2e}
\usepackage{subcaption}
\usepackage{amsmath}
\usepackage{xcolor,colortbl}
\usepackage{multirow}
\def\BibTeX{{\rm B\kern-.05em{\sc i\kern-.025em b}\kern-.08em
    T\kern-.1667em\lower.7ex\hbox{E}\kern-.125emX}}

\definecolor{Gray}{gray}{0.82}
\definecolor{LightGray}{gray}{0.88}
\definecolor{LighterGray}{gray}{0.92}
\definecolor{LightCyan}{rgb}{0.88,1,1}
\definecolor{lightyellow}{rgb}{1,1,0.85}
\definecolor{magicmint}{rgb}{0.67, 0.94, 0.82}

\begin{document}

\title{Two-Dimensional Unknown View Tomography from
Unknown Angle Distributions}

\author{\IEEEauthorblockN{Kaishva Chintan Shah}
\IEEEauthorblockA{\textit{Department of Electrical Engineering} \\
\textit{Indian Institute of Technology Bombay}\\
Mumbai, India \\
kaishvashah@gmail.com}
\and
\IEEEauthorblockN{Karthik S. Gurumoorthy}
\IEEEauthorblockA{\textit{Walmart Global Tech} \\
Bengaluru, India \\
karthik.gurumoorthy@gmail.com}
\and
\IEEEauthorblockN{Ajit Rajwade }
\IEEEauthorblockA{\textit{Department of Computer Science and Engineering} \\
\textit{Indian Insititute of Technology Bombay}\\
Mumbai, India \\
ajitvr@cse.iitb.ac.in}
}

\maketitle

\begin{abstract}
This study presents a technique for 2D tomography under unknown viewing angles when the distribution of the viewing angles is also unknown. Unknown view tomography (UVT) is a problem encountered in cryo-electron microscopy and in the geometric calibration of CT systems. There exists a moderate-sized literature on the 2D UVT problem, but most existing 2D UVT algorithms assume knowledge of the angle distribution which is not available usually. Our proposed methodology formulates the problem as an optimization task based on cross-validation error, to estimate the angle distribution jointly with the underlying 2D structure in an alternating fashion. We explore the algorithm's capabilities for the case of two probability distribution models: a semi-parametric mixture of von Mises densities and a probability mass function model. We evaluate our algorithm's performance under noisy projections using a PCA-based denoising technique and Graph Laplacian Tomography (GLT) driven by order statistics of the estimated distribution, to ensure near-perfect ordering, and compare our algorithm to intuitive baselines.
\end{abstract}

\begin{IEEEkeywords}
tomography, unknown view tomography, order statistics, cross-validation
\end{IEEEkeywords}

\section{Introduction}
Tomography is a technique used to determine the internal structure (density map $g:\mathbb{R}^2 \rightarrow \mathbb{R}$) of an object from Radon projections $\{\boldsymbol{y_i}\}_{i=1}^N$ acquired respectively from multiple angles $\{\theta_i\}_{i=1}^N$  \cite{kak2001principles}. The Radon projections are represented as follows where $\rho$ is an offset:
\begin{equation}
y_i(\rho) = \mathcal{R}_{\theta_i}g(\rho) = \int \int_{\mathbb{R}^2} g(x,y) \delta(\rho - x \cos \theta_i - y \sin \theta_i) dx dy. 
\end{equation}
If $\boldsymbol{g}$ is represented by a discrete image of size $S \times S$, then each $\boldsymbol{y_i}$ is a 1D vector containing $S$ elements. 

Tomographic imaging is widely used in various fields, including medicine (particularly computed tomography--CT, or positron emission tomography--PET), materials science, and structural biology \cite{kak2001principles}. 
%In particular, tomography is essential in Cryo-Electron Microscopy (CryoEM), where it helps to reconstruct three-dimensional (3D) structures of biological macromolecules from a series of two-dimensional (2D) projections captured at different orientations.
% Refer the para to the experiments figure
Unlike computed tomography (CT), where the projection angles are known in advance, there exist tomographic reconstruction applications such as cryo-electron microscopy (CryoEM) which face the challenge of reconstructing 3D structures from projections taken under unknown viewing angles that follow a potentially non-uniform distribution \cite{herman2014computational}. The challenge of unknown viewing angles also appears in phantomless calibration of CT imaging systems \cite{meng2012online}.

The 2D regime of UVT has been studied previously in \cite{basu2000feasibility,basu2000uniqueness,phan2017moment,Coifman2008,Singer2013,fang2010estimating}. In some papers, nonlinear dimensionality reduction algorithms such as Laplacian eigenmaps \cite{Coifman2008} or spherical multidimensional scaling \cite{fang2010estimating,Fang2011} have been employed to provide an angular ordering for the available (noisy) 1D projections. After the ordering step, the concept of order statistics is used for performing angle assignment. However, these algorithms assume prior knowledge of the angle distribution (mostly taken to be $\text{Uniform}(0,2\pi)$) for reconstruction, although in most applications, the distribution is almost always unknown and is often non-uniform. Moment-based methods for tomography have been employed in \cite{basu2000feasibility,basu2000uniqueness,phan2017moment,malhotra2016tomographic} for simultaneous angle and structure determination. Apart from \cite{phan2017moment}, these do not require knowledge of the angle distribution, but such geometric moments are not resilient to noise. The method in \cite{phan2017moment} assumes the angles to be uniformly distributed. 

In this paper, we propose a novel cross-validation based approach for the 2D UVT problem to determine the angle distribution along with the 2D structure $\boldsymbol{g}$ from noisy, unordered 1D projections at unknown angles. Our method uses alternating minimization: (\textit{i}) finding the density map $g$ starting from angles assigned using order statistics of a current estimate of the distribution, initialized as $\text{Uniform}(0,2\pi)$; and (\textit{ii}) finding angle distribution (represented in this work semi-parametrically via probability mixture models, or non-parametrically via probability mass functions), so as to minimize a carefully defined cross-validation error (CVE) using the current estimate of the image. The CVE is known to be a purely data-driven proxy for the unobservable mean squared error. It has been widely used for model selection in machine learning and compressed sensing \cite{zhang2014theoretical}, but its application here to obtain angle distributions in the 2D UVT problem is novel and useful.  

%Given the critical role of projection ordering in our method, we enhance its robustness by incorporating a PCA-based denoising technique \cite{Singer2013} and Graph Laplacian Tomography (GLT) \cite{Coifman2008}, which helps to maintain near-perfect ordering even under noisy conditions. %Additionally, we compare the reconstructed image from our algorithm with the ideal reconstructed image, assuming the angles were known, to evaluate the effectiveness of our method. The evaluation is performed using mean squared error (MSE), peak signal-to-noise ratio (PSNR), and correlation coefficient (CC).
The recent work from \cite{zehni2022adversarial} employs generative adversarial networks (GANs) to jointly recover the image and the angle distribution by matching the empirical \emph{distribution} of the measurements with that of the generated data. However this approach requires updating the weights of a complicated network architecture, naturally involving tuning many parameters for the choice of network architecture. Unlike this, our approach is based on the simpler concept of cross-validation. 

\section{Method}
\noindent\textbf{Order Statistics:}
Let $X_1, X_2, \ldots, X_N$ be a set of independent and identically distributed (i.i.d.) random variables with cumulative distribution function (CDF) $F(x)$ and probability density function (PDF) $f(x)$. The $k$-th order statistic, denoted by $X_{(k)}$, is the $k$-th smallest value among the random variables. The PDF $f_{(k)}(x)$ and CDF $F_{(k)}(x)$ of $X_{(k)}$ are respectively given by
% \begin{eqnarray}
%     f_{(k)}(x) = \frac{n!}{(k-1)!(n-k)!} [F(x)]^{k-1} [1 - F(x)]^{n-k} f(x),
%     \label{eq:orderstat_PDF} \\
%     F_{(k)}(x) = \sum_{j=k}^{n} \binom{n}{j} [F(x)]^{j} [1 - F(x)]^{n-j}, \label{eq:orderstat_CDF}
% \end{eqnarray}  
$f_{(k)}(x) = \frac{N!}{(k-1)!(N-k)!} [F(x)]^{k-1} [1 - F(x)]^{N-k} f(x)$ and 
$F_{(k)}(x) = \sum_{j=k}^{N} \binom{N}{j} [F(x)]^{j} [1 - F(x)]^{N-j}$, where $1 \leq k \leq N$. In our algorithm for 2D UVT, the random variables of interest are the angles of projection $\{\theta_i\}_{i=1}^N$ sampled from an unknown distribution $F(.)$. 

\medskip
\noindent\textbf{Graph Laplacians for Dimensionality Reduction:} We leverage dimensionality reduction and order statistics to assign angles to projections based on their relative positions in a sorted order, in the same manner as \cite{Coifman2008}. For this, a $N \times N$ adjacency matrix $\boldsymbol{W}$ is created such that $W_{ij}$ is a similarity measure between $\boldsymbol{y_i}$ and $\boldsymbol{y_j}$. From $\boldsymbol{W}$, a Laplacian matrix $\boldsymbol{L}$ is created. The vectors $\{\boldsymbol{y_i}\}_{i=1}^N$ are projected onto the eigenvectors of $\boldsymbol{L}$ corresponding to the two smallest (non-trivial) eigenvalues to yield coefficient vectors $\{\boldsymbol{\psi_i}\}_{i=1}^N$ where each $\boldsymbol{\psi_i} \in \mathbb{R}^2$. The projections are sorted using the placeholder angles $\{\beta_i\}_{i=1}^N$ where $\beta_i \triangleq \arctan (\psi_{i1}/\psi_{i2})$ is used. As noted in \cite{Coifman2008}, the $\beta_i$ values do not reflect the actual projection angles and are used only for angular sorting so that order statistics can be used for angle assignment (see below).  

\medskip
\noindent\textbf{Distribution models:}
The von Mises distribution is the analogue of the Gaussian for (hyper-)spherical data \cite{mardia2009directional}. Therefore, to represent the distribution of angle data, a mixture of von Mises distributions (MVF) can be used as a semi-parameteric model (as opposed to Gaussian mixture models, which are considered universal density estimators for data from $\mathbb{R}^d$). The MVF combines multiple von Mises distributions with different mean directions and concentration parameters $\{\mu_l, \kappa_l\}_{l=1}^L$ in the form
$f(\theta | \{w_l, \mu_l, \kappa_l\}_{l=1}^L) = \sum_{l=1}^{L} w_l \cdot \frac{\exp{(\kappa_l \cos{(\theta - \mu_l)})}}{2\pi I_0(\kappa_l)}$, where $w_l$ is the mixing coefficient of the $l$th component and $I_0(.)$ is the modified order-zero Bessel function of the 1st kind. Apart from the MVF, a probability mass function (PMF) model for discrete angle representation, which is defined by a set of probabilities $\{p_l\}_{l=1}^L$, such that $\forall l, p_l \in [0,1]; \sum_{l=1}^L p_l = 1$, can also be used. We denote the set of distribution parameters by $\Gamma$ -- in case of the MVF, we have $\Gamma \triangleq \{w_l,\mu_l,\kappa_l\}_{l=1}^L$, whereas for a PMF, we have $\Gamma \triangleq \{p_l\}_{l=1}^L$.

\medskip
\noindent\textbf{Algorithm:}
Our procedure for 2D UVT is summarized in Alg.~\ref{alg:mvf_estimation} for the MVF model. The relevant changes for the PMF model will be mentioned later. 
The procedure is initiated by sorting the projections $\{\boldsymbol{y_i}\}_{i=1}^N$ using the Laplacian Eigenmaps (LE) technique described earlier \cite{Coifman2008}, in angular order. Let the sorted projections be denoted by $\{\boldsymbol{y_{\pi(i)}}\}_{i=1}^N$ where $\pi(.)$ denotes the (sorted) permutation of $\{1,2,...,N\}$. To each $k$th projection in this sorted sequence, we then assign angles based on the $k$th order statistics of the current estimate of the angle distribution, randomly initialized in the first iteration. 
That is, the estimated angle $\widehat{\theta}_{\pi(k)}$ of projection $\boldsymbol{y_{\pi(k)}}$ is assigned to be
$\widehat{\theta}_{\pi(k)} \triangleq E_{\Gamma}(\theta_{(k)})$, i.e. the expected value of the $k$th order statistic of the current estimate of the distribution parameterized by $\Gamma$. This expected value is computed numerically using the expression for its PDF $f_{(k)}(.)$. 

% The distribution model provides \( f(\theta) \) and \( F(\theta) \), which are used to compute the k-th order statistic probability density function (PDF) \eqref{eq:orderstat_PDF}. This PDF is used to find the expected values of the k-th order statistic, which are the angles assigned to the projections.
The sorted projections and their assigned angles form the set $\mathcal{D}(\{\widehat{\theta}_{\pi(k)},\boldsymbol{y_{\pi(k)}}\}_{k=1}^N)$. This set $\mathcal{D}$ is then divided into two \emph{strictly disjoint} subsets: the reconstruction set $\mathcal{D}_r$, containing 80\% of the data, and the validation set $\mathcal{D}_v$, containing the remaining 20\%. We first reconstruct an image $\boldsymbol{\hat{g}}$ from $\mathcal{D}_r$ using the filtered back projection (FBP) method. We then simulate projections at the angles in $\mathcal{D}_v(.)$. The root mean squared error (RMSE) between the simulated projections generated from the reconstructed image and those in the validation set, $\mathcal{D}_v(.)$, serves as our cross-validation error (denoted by $J(.)$ in Alg.~\ref{alg:mvf_estimation}). We minimize the cross-validation error through an iterative descent process, updating the parameters $\Gamma$ of the angle distribution and projecting them onto their feasible value set until convergence. As closed-form expressions are unavailable, gradients of $J$ w.r.t. all parameters are computed numerically. To prevent overfitting, the subsets $\mathcal{D}_r$ and $\mathcal{D}_v$ are randomly sampled from $\mathcal{D}$ in each iteration.
\begin{algorithm}[H]
\caption{Image Reconstruction and Parameter Estimation for Mixture of von Mises Functions (MVF)}
\label{alg:mvf_estimation}
\SetAlgoLined
\DontPrintSemicolon
\KwIn{Projections $\{\boldsymbol{y_i}\}_{i=1}^N$}
\KwOut{Reconstructed image $\boldsymbol{\hat{g}}$ and angle distribution parameters $\Gamma \triangleq \{\hat{\mu}_l, \hat{\kappa}_l, \hat{w}_l\}_{l=1}^L$}
\BlankLine
\textbf{Initialize}: Randomly initialize $\{\hat{\mu}_{l}, \hat{\kappa}_{l}, \hat{w}_{l}\}_{l=1}^L$\\
\textbf{Ordering}: Obtain projection order $\{\pi(i)\}_{i = 1}^N$ by LE \cite{Coifman2008} \\
\While{parameters not converged}{
    \textbf{Angle Assignment}: Assign angles $\widehat{\theta}_{\pi(i)}$ to projections based on order statistics of current MVF ; 
    %$\mathcal{D}(\{\widehat{\theta}_{\pi(i)},\boldsymbol{y_{\pi(i)}}\}_{i=1}^N)$ \\
    \textbf{Data Split}: Divide $\mathcal{D}$ into $\mathcal{D}_r$ (80\%) and $\mathcal{D}_v$ (20\%) \\
    \textbf{Reconstruction}: Compute image $\boldsymbol{\hat{g}}$ using FBP on $\mathcal{D}_r$ \\
    \textbf{Cross Validation Error}:
    $J(\{\hat{\mu}_{l}, \hat{\kappa}_{l}, \hat{w}_{l}\}_{l=1}^L) = \sum_{(\boldsymbol{y_{\pi(j)}}, \widehat{\theta}_{\pi(j)}) \in \mathcal{D}_v} \| \boldsymbol{y_{\pi(j)}} - \mathcal{R}(\boldsymbol{\hat{g}}, \widehat{\theta}_{\pi(j)}) \|_2$,
    where $\mathcal{R}$ is the Radon transform \\
    \textbf{Gradient Computation}: Calculate $\nabla_{\hat{\mu}_{k}} J$, $\nabla_{\hat{\kappa}_{k}} J$, $\nabla_{\hat{w}_{k}} J$ \\
    \textbf{Parameter Updates}:
    \begin{align*}
        \hat{w}^{\prime}_{k} &= \frac{\hat{w}_{k} \exp(-\alpha \nabla_{\hat{w}_k} J)}{\sum_{l=1}^L \hat{w}_{l} \exp(-\alpha \nabla_{\hat{w}_l} J)} \\
        \hat{\mu}^{\prime}_{k} &= \hat{\mu}_{k} - \alpha \nabla_{\hat{\mu}_k} J; 
        \hat{\kappa}^{\prime}_{k} = \hat{\kappa}_{k} - \alpha \nabla_{\hat{\kappa}_k} J\\ 
        \hat{w}_{k} &= \hat{w}^{\prime}_{k}, \hat{\kappa}_{k} = \hat{\kappa}^{\prime}_{k},
        \hat{\mu}_{k} = \hat{\mu}^{\prime}_{k}
    \end{align*}
    \textbf{Projection}: Project updated parameters to lie within valid bounds
}
\textbf{Final Reconstruction}: Reconstruct $\boldsymbol{\hat{g}}$ using $\mathcal{D}(\{\widehat{\theta}_{\pi(i)},\boldsymbol{y_{\pi(i)}}\}_{i=1}^N)$ and $\{\hat{\mu}_l,\hat{\kappa}_l, \hat{w}_l\}_{l=1}^L$.
\end{algorithm}
The procedure using the PMF model is similar to the one presented in Alg.~\ref{alg:mvf_estimation} except that the parameter updates step now updates the values of $\{p_l\}_{l=1}^L$ via mirror descent in the form: $\hat{p}_{k}^{\prime} = \frac{\hat{p}_{k} \exp(-\alpha \nabla_{\hat{p}_k} J)}{\sum_{l=1}^K \hat{p}_{l} \exp(-\alpha \nabla_{\hat{p}_l} J)}; \hat{p}_k = \hat{p}_{k}^{\prime}$. 
% \begin{algorithm}[H]
% \caption{Parameter Estimation for Probability Mass Function (PMF) Distribution}
% \label{alg:pmf_estimation}
% \SetAlgoLined
% \DontPrintSemicolon
% \KwIn{Projections $\{y_i\}_{i=1}^N$}
% \KwOut{Reconstructed image $g$ and estimated probabilities $\{\hat{p}_k\}_{k=1}^K$}
% \BlankLine
% \textbf{Initialize}: Randomly set $\{\hat{p}_{k,1}\}_{k=1}^K$ with $\sum_{k=1}^K \hat{p}_{k,1} = 1$\\
% \textbf{Ordering}: Obtain projection order $\{\pi_i\}$ using GLT \\
% \While{parameters not converged}{
%     \textbf{Angle Assignment}: Assign angles $\{\theta_i\}$ to projections based on order statistics of the current PMF; $\mathcal{D(\hat{\theta}, P)}$\\
%     \textbf{Data Split}: Divide $\mathcal{D}$ into $\mathcal{D}_r$ (80\%) and $\mathcal{D}_v$ (20\%) \\
%     \textbf{Reconstruction}: Compute image $g$ using FBP on $\mathcal{D}_r$ \\
%     \textbf{Cross Validation Error}:
%     \[
%     J() = \sum_{(P_j, \hat{\theta_j}) \in \mathcal{D}_v} \| P_j - \mathcal{R}(g, \hat{\theta_j}) \|^2,
%     \]
%     where $\mathcal{R}$ is the Radon transform \\
%     \textbf{Gradient Computation}: Calculate $\nabla_{\hat{p}_k} L$ for each bin $k$ \\
%     \textbf{Probability Update} (Mirror Descent update):
%     \[
%     \hat{p}_{k,n+1} = \frac{\hat{p}_{k,n} \exp(-\alpha \nabla_{\hat{p}_k} L)}{\sum_{l=1}^K \hat{p}_{l,n} \exp(-\alpha \nabla_{\hat{p}_l} L)} 
%     \]
%     % \textbf{Projection}: Ensure updated probabilities satisfy $\sum_{k=1}^K \hat{p}_k = 1$ and $\hat{p}_k \geq 0$ \\
% }
% \end{algorithm}
\begin{figure*}
    \centering
    \includegraphics[scale = 0.74, width = 0.84\textwidth]
    {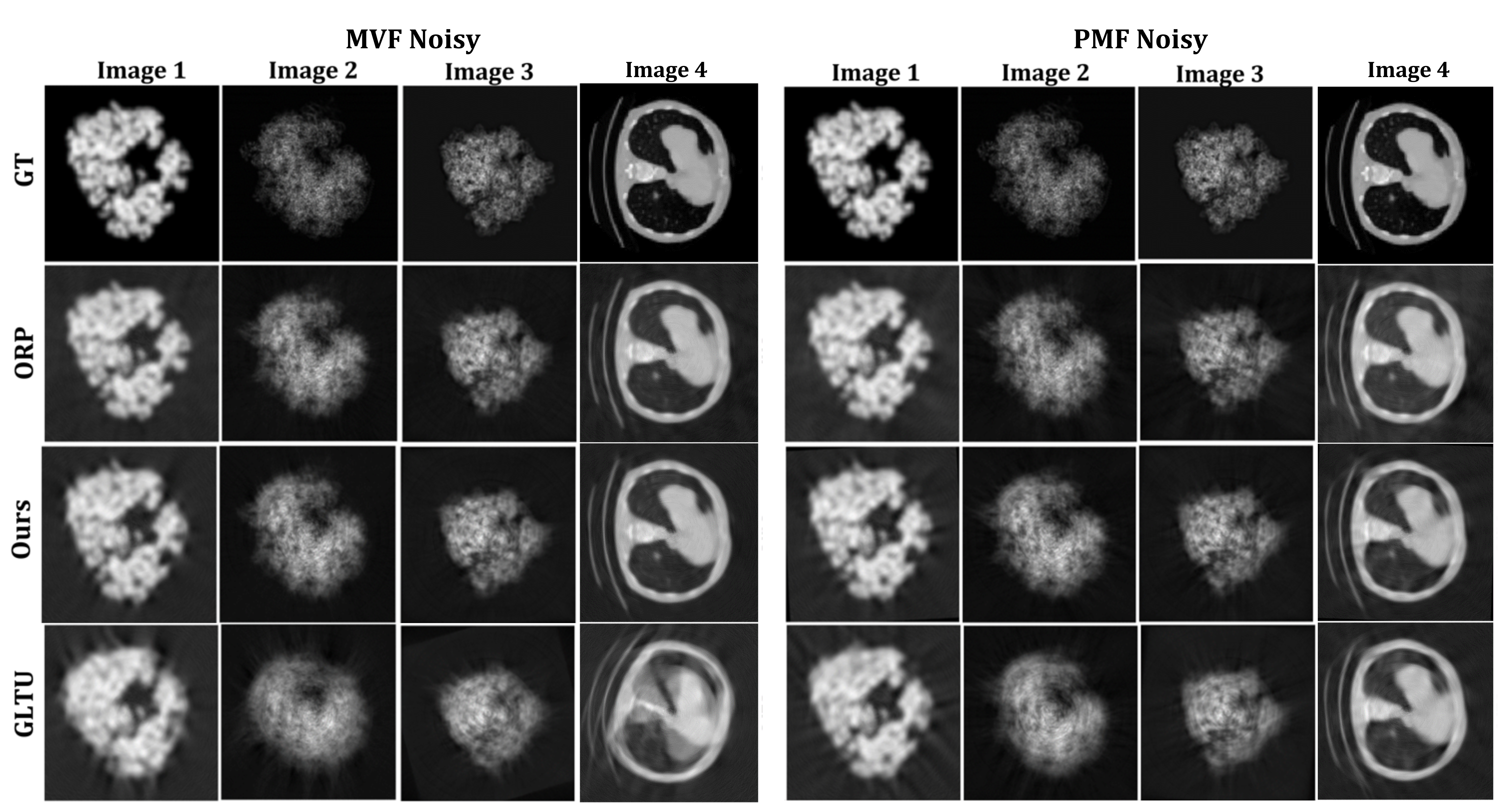}
    \caption{Visual comparison between the ground truth and reconstructions from 5000 noisy projections, obtained from our algorithm, ORP, and graph Laplacian assuming uniform distribution (GLTU) as in \cite{Coifman2008}. Our technique where we learn the distribution, is able to reconstruct image details much better than GLTU.}
    \label{fig:Recons_noisy}
\end{figure*}
The time complexity per-iteration is: $ O(d^2) + O(|D_v| \cdot d^{1.5}) + O(|D_v| \cdot \sqrt{d}) + O(L), $
where $O(d^2)$ is the time for FBP ($d$ = number of image pixels), $O(|D_v| \cdot d^{1.5})$ is the time for CVE ($|D_v|$ = number of validation angles, each projection = $\sqrt{d}$ pixels), $O(|D_v| \cdot \sqrt{d})$ is the time for gradient computation, and $O(L)$ is the time for parameter updates ($L$ = number of parameters).
\section{Experiments}
We conducted experiments on four different images of size $512 \times 512$. Three of the images are 2D slices extracted from 3D reconstructed maps of three well-known macromolecules--80S ribosome from \cite{khatter2015structure} (Image 1), a spore ribosome structure from \cite{mclaren2023cryoem} (Image 2) and a 70S E. Coli ribosome \cite{beckert2018structure} (Image 3). The fourth image is a lung CT scan \cite{Jin2017}. For each image, we generated projections at angles randomly sampled from a distribution following the PMF and MVF models. Noise (i.i.d.) from $\mathcal{N}(0,\sigma^2)$ was added to the projections with $\sigma \triangleq 0.15 \times$ the average absolute value of the (clean) projections. The results of our method were compared to those of the baseline technique GLTU (Graph Laplacian Tomography with Uniformly Spaced Angles), which orders projections using Laplacian eigenmaps and assumes a uniform distribution for the angles \cite{Coifman2008}. The results were also compared to those with an oracular reconstruction procedure (ORP) that uses the original projection angles with FBP, and acts as an `upper baseline'. For all methods, image reconstruction was preceded by a denoising step using PCA, similar to the one in \cite{Singer2013}. We were unable to numerically compare our technique to \cite{zehni2022adversarial} due to unavailability of their implementation. Results of all methods were compared using RRMSE, correlation coefficient (CC) and structural similarity index (SSIM). These are computed as follows: 
${RRMSE} = \frac{\|\boldsymbol{\hat{g}} - \boldsymbol{g}\|_2}{\|\boldsymbol{g}\|_2}; \text{CC} = \frac{\sum_{i}(\hat{g}_i - \mu_{\hat{g}})(g_i - \mu_{g})}{\sqrt{\sum_{i}(\hat{g}_i - \mu_{\hat{g}})^2 \sum_{i}(g_i - \mu_{g})^2}} \label{eq:cc};\text{SSIM}(\boldsymbol{\hat{g}}, \boldsymbol{g}) = \frac{(2\mu_{\hat{g}}\mu_{g} + C_1)(2\sigma_{\hat{g}g} + C_2)}{(\mu_{\hat{g}}^2 + \mu_{g}^2 + C_1)(\sigma_{\hat{g}}^2 + \sigma_{g}^2 + C_2)} $
where $\boldsymbol{\hat{g}}$ is the estimate of the original image $\boldsymbol{g}$,
$\hat{g}_i$ and $g_i$ are pixel values of $\boldsymbol{\hat{g}}$ and $\boldsymbol{g}$,
$\mu_{\hat{g}}$ and $\mu_{g}$ are the means of $\boldsymbol{\hat{g}}$ and $\boldsymbol{g}$,
$\sigma_{\hat{g}}^2$ and $\sigma_{g}^2$ are the variances of $\boldsymbol{\hat{g}}$ and $\boldsymbol{g}$,
$\sigma_{\hat{g}g}$ is the covariance between $\boldsymbol{\hat{g}}$ and $\boldsymbol{g}$,
$C_1$ and $C_2$ are constants to stabilize the division.

\begin{figure*}[h]
    \centering
    \includegraphics[scale = 0.75, width = 0.94\textwidth]{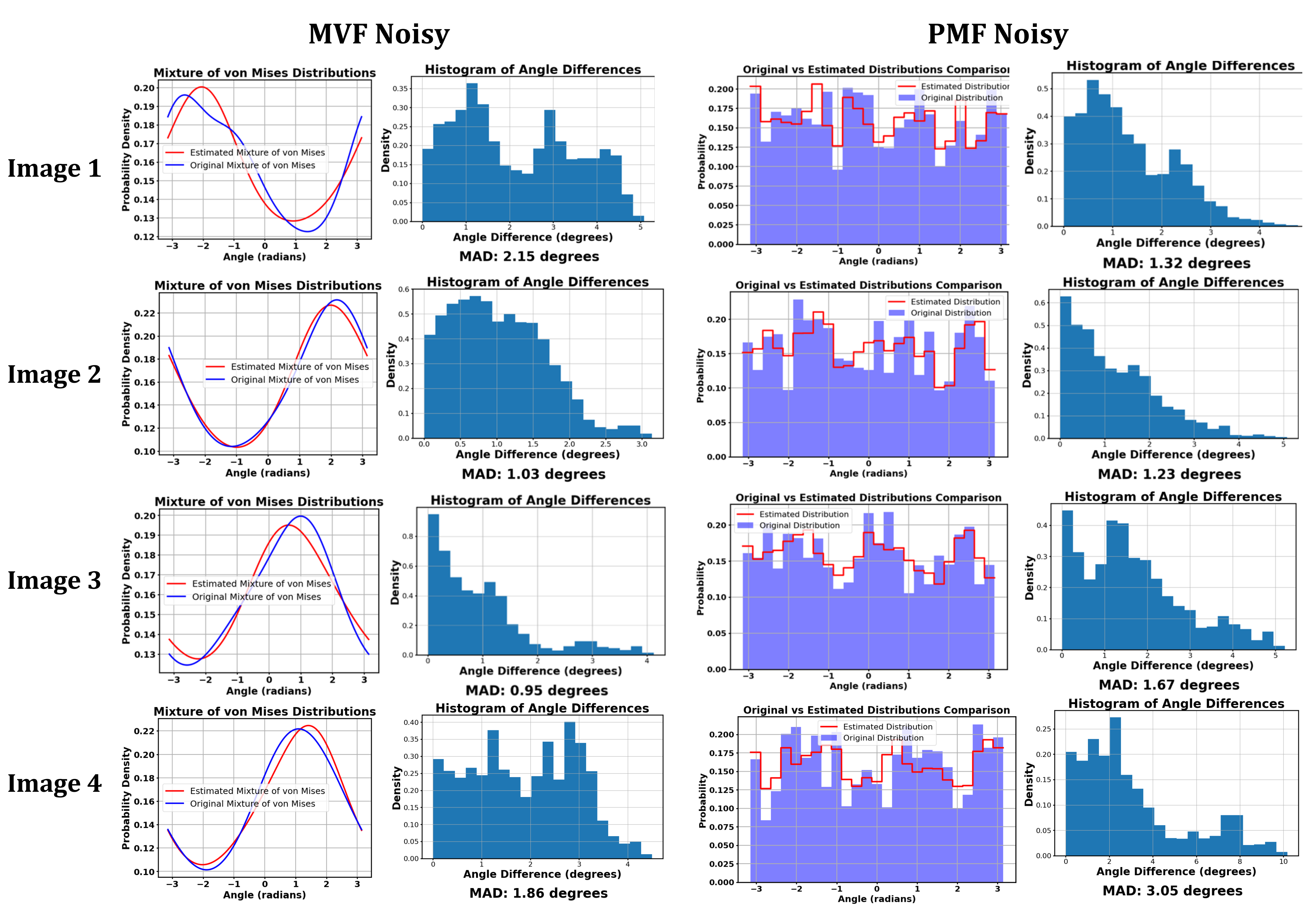}
    \caption{Comparison of the estimated and original distributions for MVF (col. 1) and PMF models (col. 3), along with the histogram of absolute angle differences (cols. 2 and 4 resp.). The estimated and original MVF both contain 5 components. The estimated PMF was computed with 25 bins, whereas the original PMF, which was generated using 50 bins, was combined into 25 bins for representation purposes.}
    \label{fig:Dist_noisy}
\end{figure*}

\begin{figure*}
    \centering
    \includegraphics[width=0.9\linewidth]{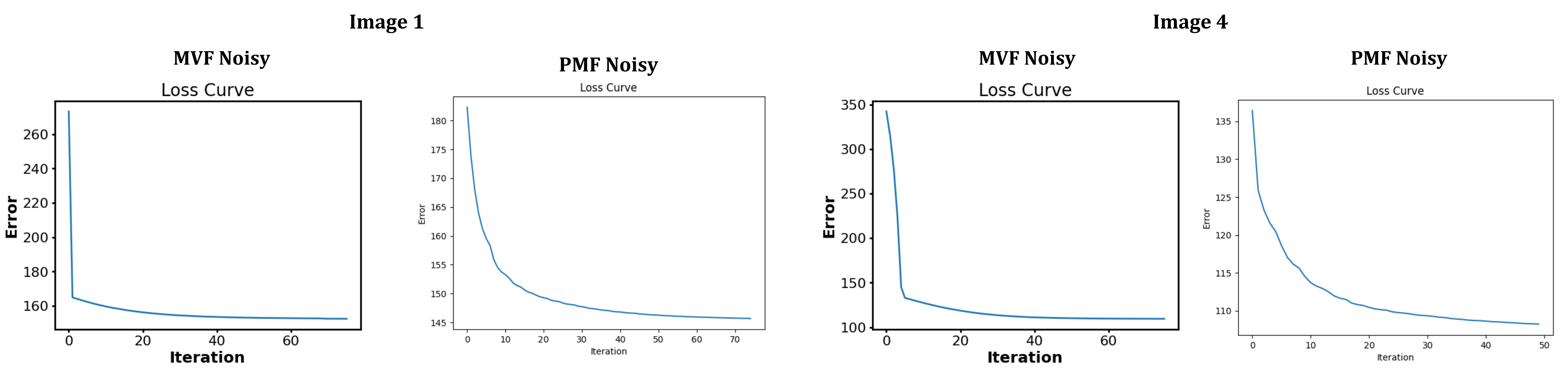}
    \caption{The cross-validation error is shown against iterations for images 1 and 4 and both distribution models (PMF and MVF).}
    \label{fig:convergence_plots}
\end{figure*}

The RRMSE, CC and SSIM were also computed between the output of our technique and that of ORP since it is an oracular baseline. Assuming knowledge of the first oracular-ordered projection (since UVT estimates can be computed only upto an unknown rotation \cite{basu2000uniqueness}), we also calculated the mean absolute difference (MAD) between the original angles and those estimated by our algorithm, measured in degrees. As shown in Fig. \ref{fig:Dist_noisy}, we compare the estimated and original distributions, along with the histogram of absolute differences between the estimated and ground truth angles. Fig. \ref{fig:Recons_noisy} presents a visual comparison of reconstructions from ORP, our method, and GLTU. The performance metrics in Table \ref{tab:Metrics_Combined} as well as the visual results in Fig. \ref{fig:Recons_noisy} demonstrate that our method clearly outperforms GLTU. The convergence plots of images 1 and 4 are shown for both distribution models (MVF and PMF) \ref{fig:convergence_plots}.
% \begin{center}
% \begin{gather}
% \text{RRMSE} = \frac{\|\boldsymbol{\hat{g}} - \boldsymbol{g}\|_2}{\|\boldsymbol{g}\|_2}
% \label{eq:rrmse}\\
% \text{CC} = \frac{\sum_{i}(\hat{g}_i - \mu_{\hat{g}})(g_i - \mu_{g})}{\sqrt{\sum_{i}(\hat{g}_i - \mu_{\hat{g}})^2 \sum_{i}(g_i - \mu_{g})^2}}
% \label{eq:cc}\\
% \text{SSIM}(\boldsymbol{\hat{g}}, \boldsymbol{g}) = \frac{(2\mu_{\hat{g}}\mu_{g} + C_1)(2\sigma_{\hat{g}g} + C_2)}{(\mu_{\hat{g}}^2 + \mu_{g}^2 + C_1)(\sigma_{\hat{g}}^2 + \sigma_{g}^2 + C_2)}
% \label{eq:ssim}
% \end{gather}
% \end{center}

\begin{table*}[h]
    \centering
    \small % Reduced font size for compactness
    \renewcommand{\arraystretch}{1.2} % Adjust row height for better readability
    \begin{tabular}{|l|c|c|c|c|c|c|}
        \hline
        \cellcolor{LightCyan}\textbf{Image} & \cellcolor{LightCyan}\textbf{Model} & \cellcolor{LightCyan}\textbf{Our vs. GT} & \cellcolor{LightCyan}\textbf{Our vs. ORP} & \cellcolor{LightCyan}\textbf{GLTU vs. GT} & \cellcolor{LightCyan}\textbf{GLTU vs. ORP} & \cellcolor{LightCyan}\textbf{ORP vs. GT} \\
        \hline
        \multirow{2}{*}{\textbf{Image 1}} & MVF & 0.34/0.98/0.28 & 0.12/0.98/0.65 & 0.43/0.91/0.19 & 0.25/0.91/0.49 & 0.33/0.99/0.33 \\
         & PMF & 0.31/0.99/0.33 & 0.12/0.99/0.71 & 0.34/0.98/0.29 & 0.13/0.98/0.65 & 0.35/0.99/0.32 \\
        \hline
        \multirow{2}{*}{\textbf{Image 2}} & MVF & 0.41/0.97/0.41 & 0.10/0.99/0.84 & 0.51/0.88/0.28 & 0.29/0.91/0.66 & 0.38/0.98/0.48 \\
         & PMF & 0.42/0.96/0.42 & 0.12/0.99/0.83 & 0.45/0.93/0.34 & 0.18/0.97/0.72 & 0.38/0.97/0.47 \\
        \hline
        \multirow{2}{*}{\textbf{Image 3}} & MVF & 0.25/0.97/0.83 & 0.08/0.99/0.90 & 0.36/0.91/0.69 & 0.23/0.95/0.73 & 0.22/0.97/0.83 \\
         & PMF & 0.25/0.96/0.79 & 0.13/0.99/0.85 & 0.31/0.93/0.73 & 0.20/0.97/0.78 & 0.21/0.97/0.83 \\
        \hline
        \multirow{2}{*}{\textbf{Image 4}} & MVF & 0.33/0.98/0.35 & 0.11/0.98/0.75 & 0.53/0.81/0.22 & 0.35/0.81/0.53 & 0.31/0.99/0.39 \\
         & PMF & 0.14/0.98/0.63 & 0.45/0.98/0.41 & 0.37/0.94/0.31 & 0.20/0.95/0.66 & 0.36/0.98/0.37 \\
        \hline
    \end{tabular}
    \caption{Comparison of measures (RRMSE $\downarrow$, CC $\uparrow$, SSIM $\uparrow$) for images using MVF and PMF models.}
    \label{tab:Metrics_Combined}
\end{table*}

\section{Discussion and Conclusion}
Unlike the GAN-based approach in \cite{zehni2022adversarial}, our approach does not use the concept of matching (via a critic or adversary) the \emph{distribution} of the observed projections with that of the projections generated from the current estimate of the image and projection angles. Instead, our technique seeks to find an angle distribution (governed by PMF or MVF parameters) in such a way that the corresponding reconstructed image (obtained from projections whose angles are determined by order statistics of the estimated distribution) is able to \emph{explain} individual projections from the \emph{unseen} validation set of projections. This is expressed mathematically by the validation error $J$ in Alg.~\ref{alg:mvf_estimation}. Our approach is also very different from the maximum likelihood approach from \cite[Equation 21]{zehni2022adversarial} because we use the reconstruction set $\mathcal{D}_r$ (for image reconstruction given an angle assignment based on the current estimate of the angle distribution) and the validation set $\mathcal{D}_v$ (for estimating the angle distribution), which are completely \emph{disjoint}. 

There are various other types of model selection criteria that have been used in image or signal reconstruction -- for example, the Morozov's discrepancy principle \cite{anzengruber2009morozov,bohra2019variance} to determine how well the reconstructed image conforms to the statistics of the measurement noise   (i.e., $\mathcal{N}(0,\sigma^2)$ in our case), or different variants of the Bayesian information criterion \cite{Gohain2023}. 
Future work could involve incorporating these criteria for angle determination in the 2D UVT problem. Likewise, our method could also be incorporated in moment-based approaches such as \cite{phan2017moment} which use angle distributions.\\
The numerical results in this paper use a PCA-based denoising technique \cite{Singer2013}, but more sophisticated denoisers such as those with CNNs may be able to further enhance  numerical performance. Recently our group has developed analytical bounds for 2DUVT reconstructions given projections from unknown angles with known distribution \cite{shah2023analysis}. Extension of these bounds to the case where the distribution is obtained from noisy projections using the technique in this paper is an important avenue for future work. 
\newpage
\bibliographystyle{IEEEbib}
\bibliography{ICASSP_2DUVT_arxv}

\begin{thebibliography}{10}

\bibitem{kak2001principles}
Avinash~C Kak and Malcolm Slaney,
\newblock {\em Principles of computerized tomographic imaging},
\newblock SIAM, 2001.

\bibitem{herman2014computational}
Gabor~T Herman and Joachim Frank,
\newblock {\em Computational methods for three-dimensional microscopy reconstruction},
\newblock Springer, 2014.

\bibitem{meng2012online}
Yuanzheng Meng, Hui Gong, and Xiaoquan Yang,
\newblock ``Online geometric calibration of cone-beam computed tomography for arbitrary imaging objects,''
\newblock {\em IEEE transactions on medical imaging}, vol. 32, no. 2, pp. 278--288, 2012.

\bibitem{basu2000feasibility}
Samit Basu and Yoram Bresler,
\newblock ``Feasibility of tomography with unknown view angles,''
\newblock {\em IEEE Transactions on Image Processing}, vol. 9, no. 6, pp. 1107--1122, 2000.

\bibitem{basu2000uniqueness}
Samit Basu and Yoram Bresler,
\newblock ``Uniqueness of tomography with unknown view angles,''
\newblock {\em IEEE transactions on image processing}, vol. 9, no. 6, pp. 1094--1106, 2000.

\bibitem{phan2017moment}
Minh~Son Phan, {\'E}tienne Baudrier, Lo{\"\i}c Mazo, and Mohamed Tajine,
\newblock ``Moment-based angular difference estimation between two tomographic projections in 2{D} and 3{D},''
\newblock {\em Journal of Mathematical Imaging and Vision}, vol. 57, pp. 164--182, 2017.

\bibitem{Coifman2008}
Ronald Coifman, Yoel Shkolnisky, Fred Sigworth, and Amit Singer,
\newblock ``Graph laplacian tomography from unknown random projections,''
\newblock {\em IEEE Transactions on Image Processing}, vol. 17, pp. 1891--1899, 2008.

\bibitem{Singer2013}
A~Singer and H-T Wu,
\newblock ``Two-dimensional tomography from noisy projections taken at unknown random directions,''
\newblock {\em SIAM journal on imaging sciences}, vol. 6, no. 1, pp. 136--175, 2013.

\bibitem{fang2010estimating}
Yi~Fang, Sundar Murugappan, and Karthik Ramani,
\newblock ``Estimating view parameters from random projections for tomography using spherical mds,''
\newblock {\em BMC medical imaging}, vol. 10, pp. 1--9, 2010.

\bibitem{Fang2011}
S.~V. N.~Vishwanathan Y.~Fang, M.~Sun and K.~Ramani,
\newblock ``s{LLE}: Spherical locally linear embedding with applications to tomography,''
\newblock in {\em CVPR}, 2011.

\bibitem{malhotra2016tomographic}
Eeshan Malhotra and Ajit Rajwade,
\newblock ``Tomographic reconstruction from projections with unknown view angles exploiting moment-based relationships,''
\newblock in {\em 2016 IEEE International Conference on Image Processing (ICIP)}. IEEE, 2016, pp. 1759--1763.

\bibitem{zhang2014theoretical}
Jinye Zhang, Laming Chen, Petros~T Boufounos, and Yuantao Gu,
\newblock ``On the theoretical analysis of cross validation in compressive sensing,''
\newblock in {\em IEEE International Conference on Acoustics, Speech and Signal Processing (ICASSP)}. IEEE, 2014, pp. 3370--3374.

\bibitem{zehni2022adversarial}
Mona Zehni and Zhizhen Zhao,
\newblock ``An adversarial learning based approach for 2{D} unknown view tomography,''
\newblock {\em IEEE Transactions on Computational Imaging}, vol. 8, pp. 705--720, 2022.

\bibitem{mardia2009directional}
Kanti~V Mardia and Peter~E Jupp,
\newblock {\em Directional statistics},
\newblock John Wiley \& Sons, 2009.

\bibitem{khatter2015structure}
Heena Khatter, Alexander~G Myasnikov, S~Kundhavai Natchiar, and Bruno~P Klaholz,
\newblock ``Structure of the human 80s ribosome,''
\newblock {\em Nature}, vol. 520, no. 7549, pp. 640--645, 2015.

\bibitem{mclaren2023cryoem}
Mathew McLaren, Rebecca Conners, Michail~N Isupov, Patricia Gil-D{\'\i}ez, Lavinia Gambelli, Vicki~AM Gold, Andreas Walter, Sean~R Connell, Bryony Williams, and Bertram Daum,
\newblock ``Cryo{EM} reveals that ribosomes in microsporidian spores are locked in a dimeric hibernating state,''
\newblock {\em Nature Microbiology}, vol. 8, no. 10, pp. 1834--1845, 2023.

\bibitem{beckert2018structure}
Bertrand Beckert, Martin Turk, Andreas Czech, Otto Berninghausen, Roland Beckmann, Zoya Ignatova, J{\"u}rgen~M Plitzko, and Daniel~N Wilson,
\newblock ``Structure of a hibernating 100s ribosome reveals an inactive conformation of the ribosomal protein s1,''
\newblock {\em Nature microbiology}, vol. 3, no. 10, pp. 1115--1121, 2018.

\bibitem{Jin2017}
Kyong~Hwan Jin, Michael~T. McCann, Emmanuel Froustey, and Michael Unser,
\newblock ``Deep convolutional neural network for inverse problems in imaging,''
\newblock {\em IEEE Transactions on Image Processing}, vol. 26, no. 9, pp. 4509--4522, 2017.

\bibitem{anzengruber2009morozov}
Stephan~W Anzengruber and Ronny Ramlau,
\newblock ``Morozov's discrepancy principle for {T}ikhonov-type functionals with nonlinear operators,''
\newblock {\em Inverse Problems}, vol. 26, no. 2, pp. 025001, 2009.

\bibitem{bohra2019variance}
Pakshal Bohra, Deepak Garg, Karthik~S Gurumoorthy, and Ajit Rajwade,
\newblock ``Variance-stabilization-based compressive inversion under {P}oisson or {P}oisson--{G}aussian noise with analytical bounds,''
\newblock {\em Inverse Problems}, vol. 35, no. 10, pp. 105006, 2019.

\bibitem{Gohain2023}
Prakash~Borpatra Gohain and Magnus Jansson,
\newblock ``Robust information criterion for model selection in sparse high-dimensional linear regression models,''
\newblock {\em IEEE Transactions on Signal Processing}, vol. 71, pp. 2251--2266, 2023.

\bibitem{shah2023analysis}
Sheel Shah, Karthik~S Gurumoorthy, Kaishva Shah, and Ajit Rajwade,
\newblock ``Analysis of tomographic reconstruction of 2{D} images using the distribution of unknown projection angles,''
\newblock {\em arXiv preprint arXiv:2304.06376}, 2023.

\end{thebibliography}

\end{document}